\documentclass[10pt,twocolumn,letterpaper]{article}

\usepackage{iccv}
\usepackage{times}
\usepackage{epsfig}
\usepackage{graphicx}
\usepackage{amsmath}
\usepackage{amssymb}

\usepackage{amsmath}
\usepackage[title]{appendix}
\usepackage{bbold}
\usepackage{diagbox}
\usepackage{graphbox}
\usepackage{float}
\usepackage{footnote}
\usepackage{multirow}
\usepackage{rotating}
\usepackage{setspace}
\usepackage{sidecap}
\usepackage{tabularx}
\usepackage{mathrsfs}
\usepackage{subcaption}
\usepackage{wrapfig}
\usepackage{makecell}
\usepackage{slashbox}

\newcommand{\RR}{\mathbb{R}}

\usepackage[pagebackref=true,breaklinks=true,letterpaper=true,colorlinks,bookmarks=false]{hyperref}

\iccvfinalcopy 

\def\iccvPaperID{827} 

\ificcvfinal\fi
\begin{document}

\title{Composite Shape Modeling via Latent Space Factorization}
\author{Anastasia Dubrovina$^1$\qquad Fei Xia$^1$\qquad Panos Achlioptas$^1$\qquad Mira Shalah$^1$\\[0.3em]
Rapha\"el Groscot$^2$ \qquad Leonidas Guibas$^1$\\[0.5em]
$^1$Stanford University\qquad
$^2$PSL Research University
}

\maketitle


\begin{abstract}
  We present a novel neural network architecture, termed Decomposer-Composer, for semantic structure-aware 3D shape modeling. Our method utilizes an auto-encoder-based pipeline, and produces a novel factorized shape latent space, where the semantic structure of the shape collection translates into a data-dependent sub-space factorization, and where shape composition and decomposition become simple linear operations on the embedding coordinates. We further propose to model shape assembly using an explicit learned part deformation module, which utilizes a 3D spatial transformer network to perform an in-network volumetric grid deformation, and which allows us to train the whole system end-to-end. The resulting network allows us to perform part-level shape manipulation, unattainable by existing approaches. Our extensive ablation study, comparison to baseline methods and qualitative analysis demonstrate the improved performance of the proposed method.
\end{abstract}


\section{Introduction}
%

Understanding, modeling and manipulating 3D objects are areas of great interest to the vision and graphics communities, and have been gaining increasing popularity in recent years.
Examples of related applications include semantic segmentation~\cite{yi2016scalable}, shape synthesis ~\cite{wu2016learning, achlioptas2017representation}, 3D reconstruction~\cite{choy20163d, fan2017point}, view synthesis~\cite{xia2018gibson}, and fine-grained shape categorization~\cite{achlioptas2019shapeglot}, to name a few.
The advancement of deep learning techniques, and the creation of large-scale 3D shape datasets~\cite{chang2015shapenet} enabled researchers to learn task-specific representations directly from the existing data, and led to significant progress in all the aforementioned areas.

\begin{figure}[t]
\centering
\includegraphics[width=1\linewidth]{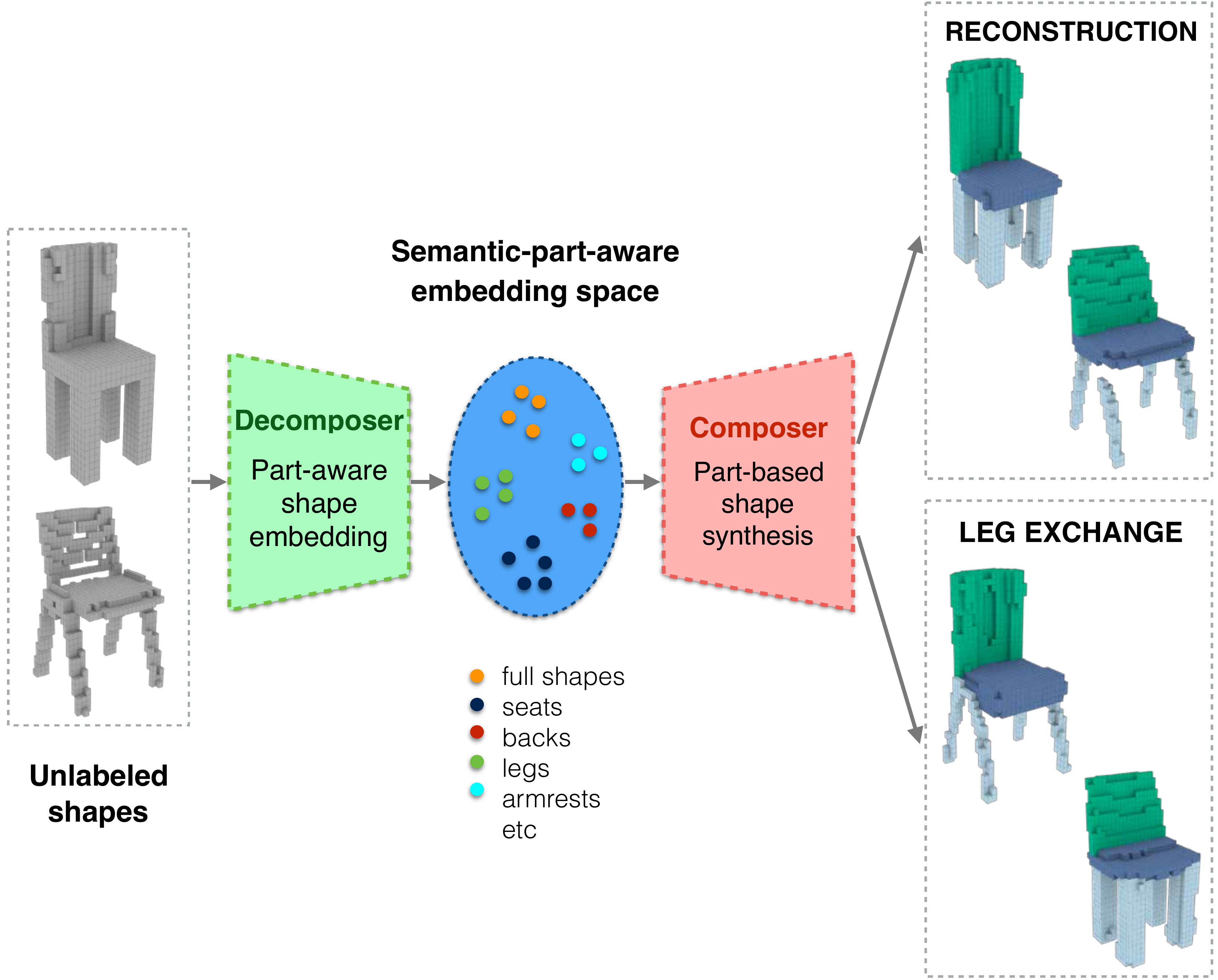}
\caption{Given unlabeled shapes, the Decomposer maps them into a factorized latent space. The Composer can either reconstruct the shapes with semantic part labels, or 
create new shapes
, for instance, by exchanging chair legs.\label{fig:teaser}}
\vspace{-1em}
\end{figure}

There is a growing interest in learning shape modeling and synthesis in a structure-aware manner, for instance, at the level of semantic shape parts.
This poses several challenges compared to approaches considering the shapes as a whole.
Semantic shape structure and shape part geometry are usually interdependent, and relations between the two must be implicitly or explicitly modeled and learned by the system.
Examples of such structure-aware shape representation-learning are \cite{nash2017shape,li2017grass,G2L18,wu2018structure}.

However, the existing approaches for shape modeling, while being part aware at the intermediate stages of the system, still ultimately operate on the low-dimensional representations of the \emph{whole} shape. For example, \cite{nash2017shape,G2L18} use a Variational Autoencoder (VAE)~\cite{kingma2013auto} to learn generative part-aware models of man-made shapes, but the latent spaces of the VAEs correspond to complete shapes, with entangled latent factors corresponding to different semantic parts.
Therefore, these and other existing approaches cannot perform part-level shape manipulation, such as single part replacement, part interpolation, or part-level shape synthesis.

Inspired by the recent efforts in image modeling to separate different image formation factors, to gain better control over image generation process and simplify editing tasks~\cite{pumarola2018ganimation,shu2018deforming,shu2017neural}, we propose a new semantic structure-aware shape modeling system.
This system utilizes an auto-encoder-based pipeline, and produces a factorized latent space which both reflects the semantic part structure of the shapes in the dataset, and compactly encodes different semantic parts' geometry. In this latent space, different semantic part embedding coordinates lie in \emph{separate linear subspaces}, and shape composition can naturally be performed by summing up part embedding coordinates. The latent space factorization is data-dependent, and is performed using learned linear projection operators. Furthermore, the proposed system operates on \emph{unlabeled} input shapes, and at test time it simultaneously infers the shape's semantic structure and compactly encodes its geometry.

Towards that end, we propose a Decomposer-Composer pipeline, schematically illustrated in Figure~\ref{fig:teaser}.
The Decomposer maps an input shape, represented by an occupancy grid, into the factorized latent space described above.
The Composer reconstructs a shape with semantic part-labels from a set of part-embedding coordinates. It explicitly learns the set of transformations to be applied to the parts, so that together they form a semantically and geometrically plausible shape. In order to learn and apply those part transformations, we employ a 3D variant of the Spatial Transformer Network (STN)~\cite{jaderberg2015spatial}. 
3D STN was previously utilized to scale and translate objects represented as 3D occupancy grids in \cite{hu2018predictive}, but to the best of our knowledge, ours is the first approach suggesting an in-network affine deformation of occupancy grids. 

Finally, to promote part-based shape manipulation, such as part replacement, part interpolation, or shape synthesis from arbitrary parts, we employ the cycle consistency constraint~\cite{zhu2017unpaired, pumarola2018ganimation, nguyen2011optimization, wang2013image}.
We utilize the fact that the Decomposer maps input shapes into a factorized embedding space, making it possible to control which parts are passed to the Composer for reconstruction. Given a batch of input shapes, we apply our Decomposer-Composer network twice, while randomly mixing part embedding coordinates before the first Composer application, and then de-mixing them into their original positions before the second Composer application. The resulting shapes are required to be as similar as possible to the original shapes, using a cycle consistency loss. 

\vspace{-1em}
\paragraph{Main contributions}
Our main contributions are: (1) A novel latent space factorization approach which enables performing shape structure manipulation using linear operations directly in the learned latent space; (2) The application of a 3D STN to perform in-network affine shape deformation, for end-to-end training and improved reconstruction accuracy; (3) The incorporation of a cycle consistency loss for improved reconstruction quality.


\begin{figure*}[t]
\centering
\includegraphics[width=1\linewidth]{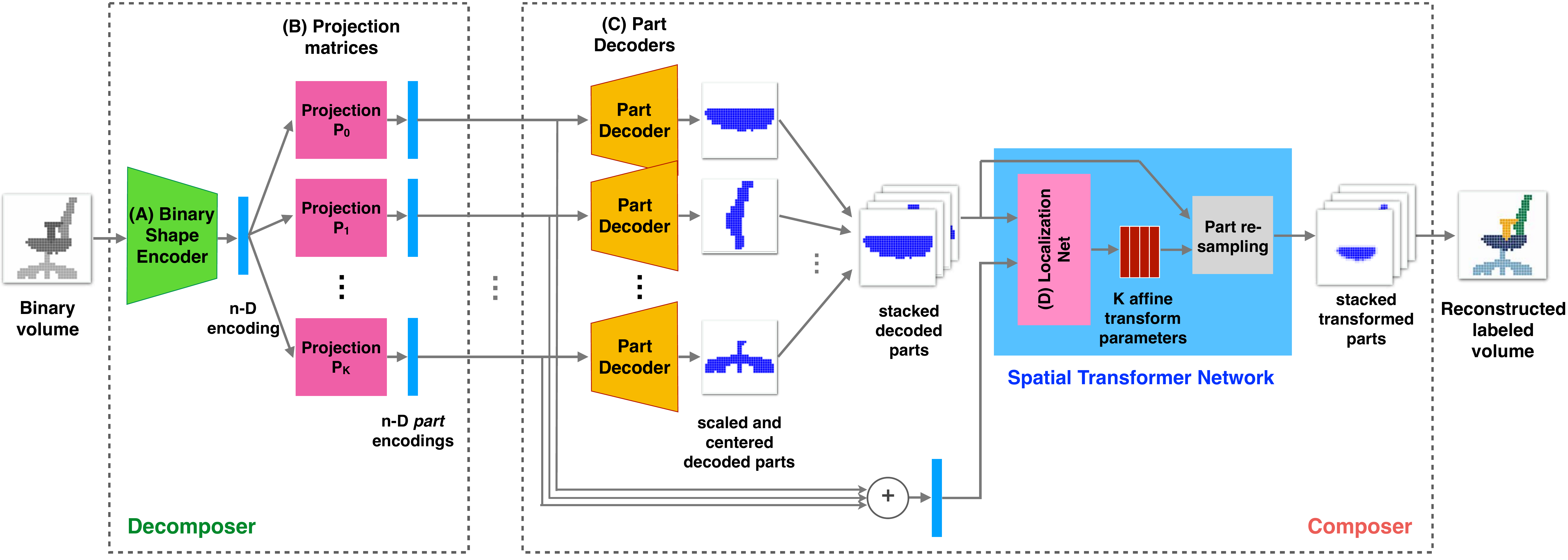}
\caption{The proposed Decomposer-Composer architecture.\label{fig:wae}}
\vspace{-1em}
\end{figure*}


\section{Related work}
\paragraph{Learning-based shape synthesis} 
Learning-based methods have been used for automatic synthesis of shapes from complex real-world domains; In a seminal work~\cite{kalogerakis2012probabilistic}, Kalogerakis \etal used a probabilistic model, which learned both continuous geometric features and discrete component structure,  for component-based shape synthesis and novel shape generation. The development of deep neural networks enabled learning high-dimensional features more easily;  3DGAN~\cite{wu2016learning} uses 3D decoders and a GAN to generate voxelized shapes. A similar approach has been applied to 3D point clouds and achieved high fidelity and diversity in shape synthesis~\cite{achlioptas2017representation}. 

Apart from generating shapes using a latent representation, some methods generate shapes from a latent representation with \emph{structure}.
SSGAN~\cite{wang2016generative} generate the shape and texture for a 3D scene in a 2-stage manner. GRASS~\cite{li2017grass} generate shapes in two stages: first, by generating orientated bounding boxes, and then a detailed geometry within those bounding boxes. Nash and Williams~\cite{nash2017shape} use use point cloud shape representation and a VAE to learn a probabilistic latent space of shapes; however, they require all training data to be in point-to-point correspondence. In a related work \cite{G2L18}, Wang \etal introduced a 3D GAN-based generative model for 3D shapes, which produced segmented and labeled into parts shapes. Unlike the latter approach, our network does not use predefined subspaces for part embedding, but learns to project the latent code of the entire shape to the subspaces corresponding to codes of different parts. 

In concurrent efforts, several deep architectures for part based shape synthesis were proposed  \cite{schor2018learning,li2019learning,pageSAGnet19,mo2019structurenet}. Schor \etal \cite{schor2018learning} 
utilized point-base shape representation, while operating on input models with known per-point parts labels. Li \etal \cite{li2019learning} and \cite{pageSAGnet19} proposed two generative networks for part-based shape synthesis, operating on labeled voxelized shapes. Mo \etal \cite{mo2019structurenet} introduced a hierarchical graph network for learning structure-aware shape generation.

\vspace{-1em}
\paragraph{Spatial transformer networks}
Spatial transformer networks (STN)~\cite{jaderberg2015spatial} allow to easily incorporate deformations into a learning pipeline. Kurenkov \etal~\cite{kurenkov2018deformnet} retrieve a 3D model from one RGB image and generate a deformation field to modify it. Kanazawa \etal~\cite{kanazawa2016learning} model articulated or soft objects with a template shape and deformations. Lin \etal~\cite{lin2018st} use STNs iteratively, to warp a foreground onto a background, and use a GAN to constrain the composition results to the natural image manifold. Hu \etal~\cite{hu2018predictive} use a 3D STN to scale and translate objects given as volumetric grids, as a part of scene generation network. Inspired by this line of work, we incorporate an affine transformation module into our network. This way, the generation module only needs to generate normalized parts, and the deformation module transforms and assembles the parts together.

\vspace{-1em}
\paragraph{Deep latent space factorization} Several approaches suggested to learn disentangled latent spaces for image representation and manipulation. $\beta$-VAE~\cite{higgins2016beta} introduce an adjustable hyperparameter $\beta$ that balances latent channel capacity and independence constraints with reconstruction accuracy. InfoGAN~\cite{chen2016infogan} achieves the disentangling of factors by maximizing the mutual information between certain channels of latent code and image labels. Some approaches disentangle the image generation process using intrinsic decomposition, such as albedo and shading~\cite{shu2017neural}, or normalized shape and deformation grid~\cite{pumarola2018ganimation,shu2018deforming}. 
The proposed approach differs from \cite{pumarola2018ganimation,shu2018deforming, shu2017neural} in that it maps both full and partial shapes into the same low dimensional embedding space, while in \cite{pumarola2018ganimation,shu2018deforming, shu2017neural}, different components have their own separated embedding spaces. 

\vspace{-1em}
\paragraph{Projection in neural networks}
Projection is widely used in representation learning. It can be used for transformation from one domain to another domain~\cite{barnes2018projecting, poelitz2014projection, poerio2018dual}, which is useful for tasks like translation in natural language processing. For example, Senel \etal \cite{senel2018semantic} use projections to map word vectors into semantic categories. 
In this work, we use a projection layer to transform a whole shape embedding into semantic part embeddings. 

\section{Our model\label{sec:our_model}}
\subsection{Decomposer network\label{sec:decomposer}}
The Decomposer network is trained to embed unlabeled shapes 
into a factorized embedding space, reflecting the shared semantic structure of the shape collection. To allow for composite shape synthesis, the embedding space has to satisfy the following two properties: factorization consistency across input shapes, and existence of a simple shape composition operator to combine latent representations of different semantic factors.
We propose to model this embedding space $V$ as a \emph{direct sum of subspaces} $\{V_i\}_{i=1}^K$, where $K$ is the number of semantic parts, and each subspace $\{V_i\}$ corresponds to a semantic part $i$, thus satisfying the factorization consistency property. The second property is ensured by the fact that every vector $v \in V$ is given by a sum of unique $v_i \in V_i$ such that $V=V_1 \oplus ... \oplus V_k $, and part composition may be performed by part embedding summation. This also implies that the decomposition and composition operations in the embedding space are fully reversible.

A simple approach for such factorization is to split the dimensions of the $n$-dimensional embedding space into $K$ coordinate groups, each group representing a certain semantic part-embedding. In this case, the full shape embedding is a concatenation of part embeddings, an approach explored in~\cite{G2L18}. This, however, puts a hard constraint on the dimensionality of part embeddings, and thus also on the representation capacity of each part embedding subspace. Given that different semantic parts may have different geometric complexities, this factorization may be sub-optimal.

Instead, we perform a data-driven learned factorization of the embedding space into semantic subspaces. We use \emph{learned} part-specific projection matrices, denoted by $\{P_i\}_{i=1}^K \in \RR^{n \times n}$. To ensure that the aforementioned two factorization properties hold, the projection matrices must form a \emph{partition of the identity} and satisfy the following three properties
\begin{align}
(1) & \,\,\, P_i^2 = P_i, \forall i,\nonumber\\
(2) & \,\,\, P_i P_j = \mathit{0} \text{ whenever } i \neq j,\nonumber\\
(3) & \,\,\, P_1 + ... + P_K = I,
\label{eq:part_of_identity}
\end{align}
where $\mathit{0}$ and $I$ are the all-zero and the identity matrices of size $n \times n$, respectively.

In practice, we efficiently implement the projection operators using fully connected layers without added biases, with a total of $K*n^2$ variables, constrained as in Equation~\ref{eq:part_of_identity}.
The projection layers receive as input a whole shape encoding, which is produced by a 3D convolutional shape encoder. The parameters of the shape encoder and the projection layers are learned simultaneously. The resulting architecture of the Decomposer network is schematically described in Figure~\ref{fig:wae}, and a detailed description of the shape encoder and the projection layer architecture is given in the supplementary material.

\subsection{Composer network}
\label{sec:cycle_consist}
The composer network is trained to reconstruct shapes with semantic part labels from \emph{sets} of semantic part embedding coordinates. The simplest composer implementation would consist of a single decoder mirroring the whole binary shape encoder (see Figure~\ref{fig:wae}), producing a semantically labelled reconstructed output shape. Such approach was used in~\cite{G2L18}, for instance. However, this straightforward method is known to fail in reconstructing thin volumetric shape parts and other fine shape details. To address this issue, we use a different approach, where we first separately reconstruct scaled and centered shape parts, using a \emph{shared part decoder}. We then produce \emph{per-part transformation parameters} and use them to deform the parts in a coherent manner, to obtain a complete reconstructed shape.

In our model, we make the simplifying assumption that it is possible to combine a given set of parts into a plausible shape by transforming them with per-part affine transformations and translations. While the true set of transformations which produce plausible shapes is significantly larger and more complex, our experiments demonstrate that the proposed simplified model is successful at producing geometrically and visually plausible results. This in-network part transformation is implemented using a 3D spatial transformer network (STN)~\cite{jaderberg2015spatial}. It consists of a localization net, which produces a set of 12-dimensional affine transformations (including translations) for all parts, and a re-sampling unit, which transforms and places the reconstructed part volumes at their correct locations in the full shape. The SNT receives as input both the reconstructed parts from the part decoder, and the sum of part encodings, for best reconstruction results. The resulting Composer architecture is schematically described in Figure~\ref{fig:wae}; its detailed description is given in the supplementary material.

We note that the proposed approach is related to the two-stage shape synthesis approach of~\cite{li2017grass}, in which a GAN is first used to synthesize oriented bounding boxes for different parts, and then the part geometry is created per bounding box using a separate part decoder. Our approach is similar, yet it works in a reversed order. Namely, we first reconstruct part geometry, and then compute per-part affine transformation parameters, which are a 12-dimensional equivalent of the oriented part bounding boxes in~\cite{li2017grass}.
Similarly to~\cite{li2017grass}, this two stage approach 
improves the reconstruction of fine geometric details. However, unlike~\cite{li2017grass}, where the GAN and the part decoder where trained separately, in our approach the two stages belong to the same reconstruction pipeline, trained simultaneously and end-to-end. 

\subsection{Cycle consistency\label{sec:cycle_augmentation}}
\begin{figure}[tb]
\begin{center}
\includegraphics[width=0.95\linewidth]{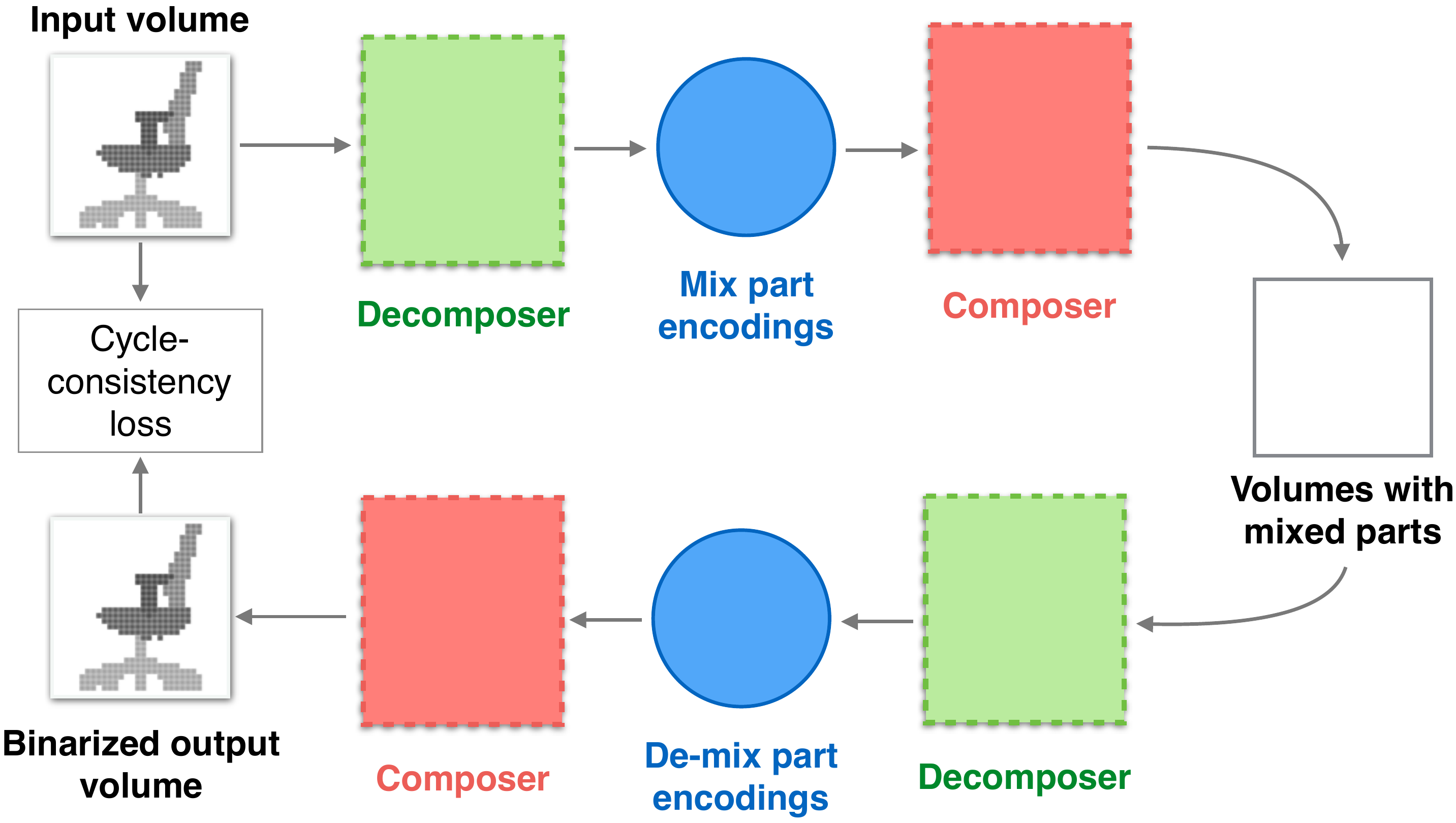}
\caption{Schematic description of the cycle consistency constraint. See Section~\ref{sec:cycle_augmentation} for details.\label{fig:cycle_consist}}
\end{center}
\vspace{-1em}
\end{figure}
Our training set is comprised of 3D shapes with ground-truth semantic part-decomposition; It does not include any training examples of synthesized composite shapes. Existing methods for such shape assembly task operate on 3D meshes with very precise segmentations, and often with additional knowledge about part connectivity \cite{xu2011photo,shen2012structure,kalogerakis2012probabilistic}. These methods cannot be applied to a dataset like ours, to produce a sufficiently large set of plausible new shapes (constructed from existing parts) to use for training a deep network for composite shape modelling. 
In order to circumvent this difficulty, and train the net to produce non-trivial part transformations for geometrically and semantically plausible part arrangements, we use a \emph{cycle consistency} constraint. It has been previously utilized in geometry processing~\cite{nguyen2011optimization}, image segmentation~\cite{wang2013image}, and more recently in neural image transformation~\cite{pumarola2018ganimation, zhu2017unpaired}.

Specifically, given a batch of $M$ training shapes $\{X\}_{i=1}^M$, we map them to the factored latent space using the Decomposer, producing $K$ semantic part encodings per input shape. We the randomly mix the part encodings of the shapes in the batch, while ensuring that after the mixing each of the new $M$ encoding sets includes exactly one embedding coordinate per semantic part. We then reconstruct the shapes with correspondingly mixed parts using the Composer. After that, these new shapes are passed to the Decomposer-Composer pipeline once again, while de-mixing part encodings produced by the second Decomposer application, to re-store the original encoding-to-shape association. The cycle consistency requirement means that the final shapes are as similar as possible to the original $M$ training shapes. We enforce it using the cycle consistency loss described in the next section. The double application of the proposed network with part encoding mixing and de-mixing is schematically described in Figure~\ref{fig:cycle_consist}.

\subsection{Loss function\label{sec:loss}}
Our loss function is defined as the following weighted sum of several loss terms
\begin{align}
\label{eq:loss}
L = & \,\, w_{\text{PI}} \mathcal{L}_{\text{PI}} + w_{\text{part}} \mathcal{L}_{\text{part}} +  w_{\text{trans}} \mathcal{L}_{\text{trans}} +  w_{\text{cycle}} \mathcal{L}_{\text{cycle}}.
\end{align}
The weights compensate for the different scales of the loss terms, and reflect their relative importance.

\vspace{-1em}
\paragraph{Partition of the identity loss} \hspace{-0.8em} $\mathcal{L}_{\text{PI}}$ measures the deviation of the predicted projection matrices from the optimal projections, as given by Equation~\ref{eq:part_of_identity}.
\begin{align}
\label{eq:pi_loss}
\mathcal{L}_{\text{proj}}(P_1,...,P_k) = & \sum_{i=1}^K \|P_i^2-P_i\|_F^2 + \sum_{\substack{i,j=1,\\i \neq j}}^K \|P_i P_j\|_F^2 + \nonumber\\ & \|P_1+...P_K - I\|_F^2.
\end{align}

\vspace{-1em}
\paragraph{Part reconstruction loss} \hspace{-0.8em} $\mathcal{L}_{\text{part}}$ is the binary cross-entropy loss between the reconstructed centered and scaled part volumes and their respective ground truth part indicator volumes, summed over $K$ parts.

\vspace{-1em}
\paragraph{Transformation parameter loss} \hspace{-0.8em} $\mathcal{L}_{\text{trans}}$ is an $L2$ regression loss between the predicted and the ground truth 12-dimensional transformation parameter vectors, summed over $K$ parts. Unlike in the original STN approach~\cite{jaderberg2015spatial}, we found that direct supervision over the transformation parameters is critical for our network convergence.

\vspace{-1em}
\paragraph{Cycle consistency loss} \hspace{-0.8em} $\mathcal{L}_{\text{cycle}}$ is a binary cross-entropy loss between ground truth input volumes and their reconstructions, obtained using two applications of the proposed network, as described in Section~\ref{sec:cycle_augmentation}.

\subsection{Training details\label{sec:training}}
The network was implemented in TensorFlow~\cite{abadi2016tensorflow}, and trained for 500 epochs with batch size 32. We used Adam optimizer~\cite{kingma2014adam} with learning rate $0.0001$, decay rate of $0.8$, and decay step size of 40 epochs.
We found it was essential to first pre-train the binary shape encoder, projection layer and part decoder parameters separately for $150$ epochs, by minimizing the part reconstruction and the partition of the identity losses and using $w_{\text{trans}}=w_{\text{cycle}}\approx 0$, for improved part reconstruction results. We then train the parameters of the spatial transformer network for another $100$ epochs, while keeping the rest of the parameters fixed. After that we resume the training with all parameters and the cycle consistency loss to fine-tune the network parameters. 
The optimal loss combination weights were empirically detected using the validation set, and set to be $w_{\text{PI}} = 0.1, w_{\text{part}} = 100, w_{\text{trans}} = 0.1, w_{\text{cycle}} = 0.1$. The network was trained on each shape category separately.



\section{Experiments}
\paragraph{Dataset}
In our experiments, we used the models from the ShapeNet 3D data collection~\cite{chang2015shapenet}, with part annotations produced by Yi \etal~\cite{yi2016scalable}. 
The shapes were converted to $32\times32\times32$ occupancy grids using binvox~\cite{nooruddin2003simplification}. Semantic part labels were first assigned to the occupied voxels according to the proximity to the labeled 3D points, and the final voxel labels were obtained using graph-cuts in the voxel domain~\cite{boykov2001fast}. 
We used the official ShapeNet train, validation and test data splits in all our experiments. Additional results for $64\times64\times64$ occupancy grids can be found in the supplementary material.



\subsection{Shape reconstruction}
\label{sec:reconstruction}
\begin{figure}[tb]
\begin{center}
    \includegraphics[width=0.15\linewidth]{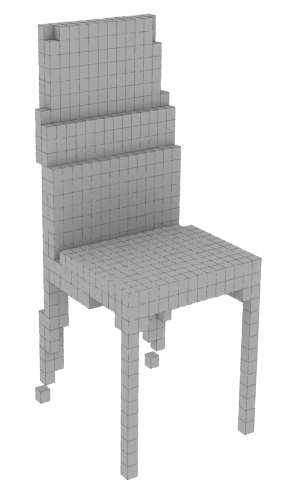}
    \hfill
    \includegraphics[width=0.15\linewidth]{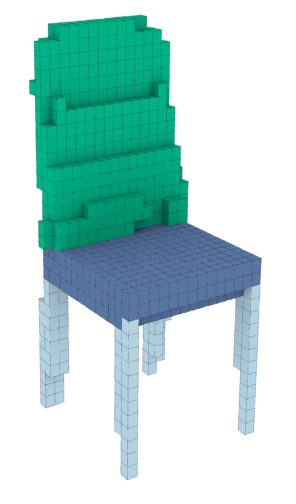}
    \hfill\hfill\hfill\hfill
    \includegraphics[width=0.15\linewidth]{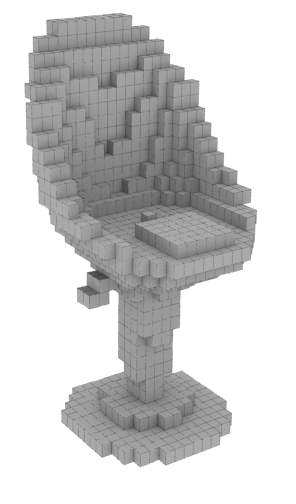}
    \hfill
    \includegraphics[width=0.15\linewidth]{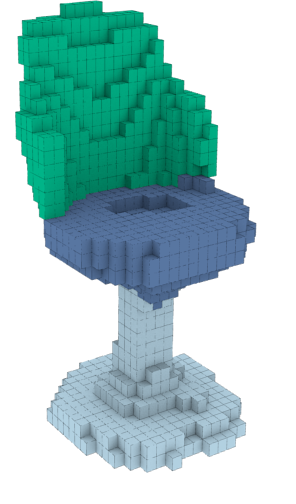}
    \hfill\hfill\hfill\hfill
    \includegraphics[width=0.15\linewidth]{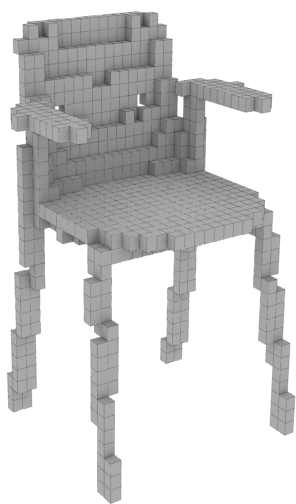}
    \hfill\hfill\hfill\hfill
    \includegraphics[width=0.15\linewidth]{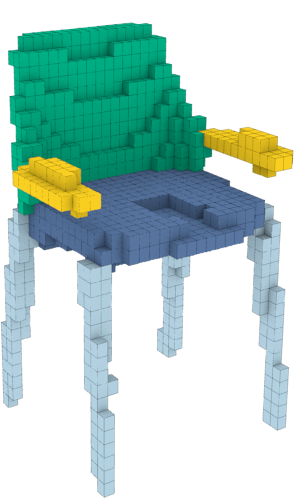}
    \\
    \includegraphics[width=0.23\linewidth]{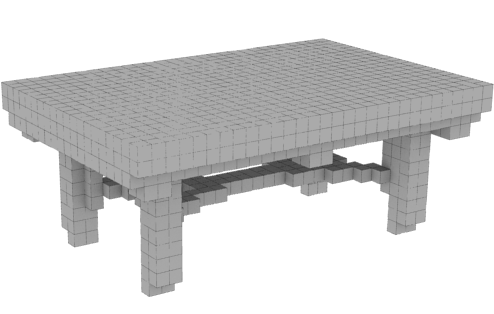}
    \hfill
    \includegraphics[width=0.23\linewidth]{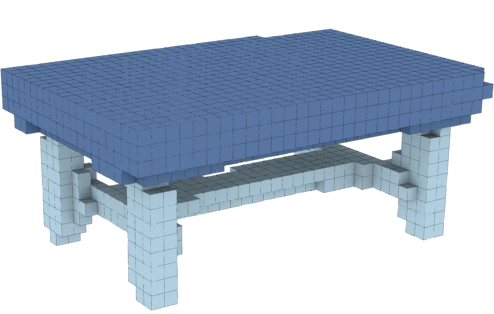}
    \hfill\hfill\hfill\hfill
    \includegraphics[width=0.23\linewidth]{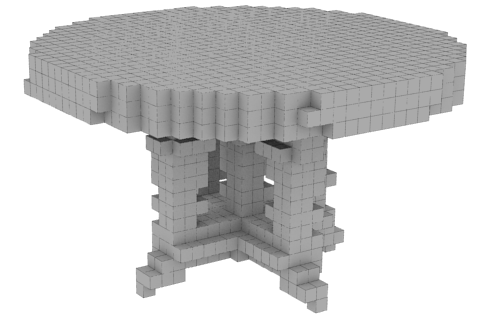}
    \hfill
    \includegraphics[width=0.23\linewidth]{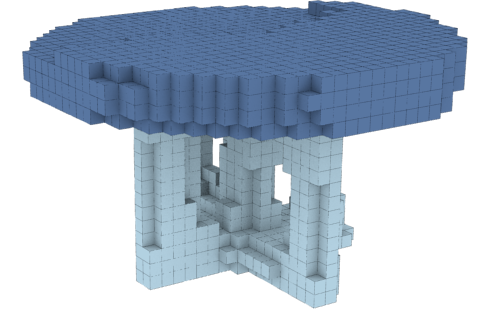}
\vspace{-0.5em}
\caption{Reconstruction results of the proposed pipeline, for chair and table shapes. Gray shapes are the input test shapes; the results are colored according to the part label.}
\label{fig:volaestn_results}
\end{center}
\vspace{-1em}
\end{figure}

Figure~\ref{fig:volaestn_results} presents the results of reconstructing semantically labeled shapes from unlabelled input shapes, using the proposed network. Note that since our method performs separate part reconstruction with part decoders and part placement with an STN, it may produce less accurate part reconstruction, as compared to segmentation approaches - for example, the handles of the reconstructed rightmost chair in Figure~\ref{fig:volaestn_results}. But, as illustrated by our quantitative study in Section~\ref{sec:ablation}, this allows us to perform better part-based shape manipulation.


\subsection{Composite shape synthesis}
\label{sec:composite_synthesis}
\begin{figure}[tb]
\begin{center}
\rotatebox{90}{\hspace{1.5em}Legs}
\hfill
\includegraphics[width=0.15\linewidth]{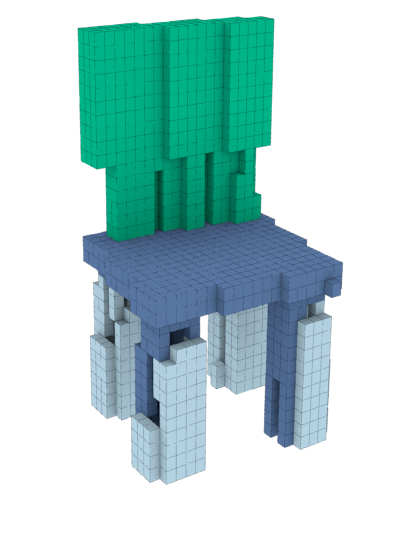}
\hfill
\includegraphics[width=0.15\linewidth]{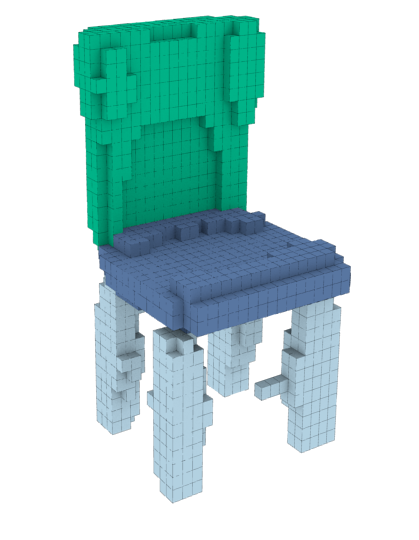}
\hfill
\includegraphics[width=0.15\linewidth]{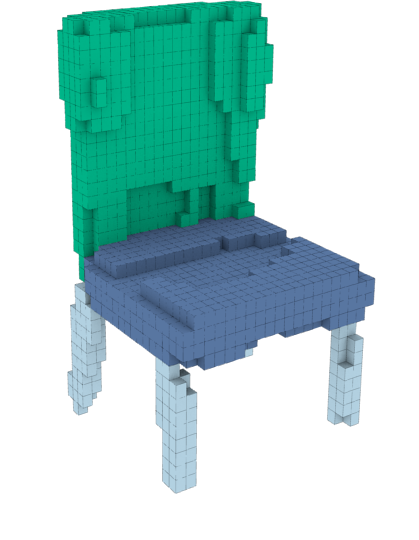}
\hfill\hfill
\includegraphics[width=0.15\linewidth]{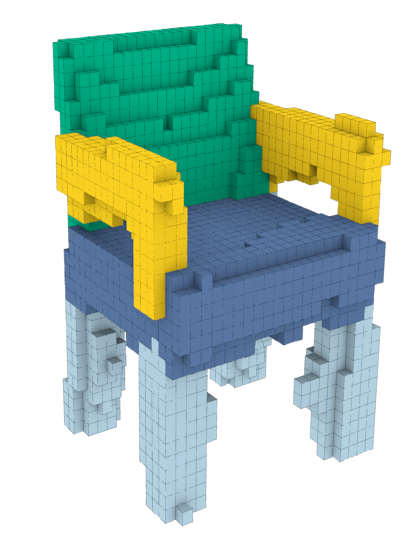}
\hfill
\includegraphics[width=0.15\linewidth]{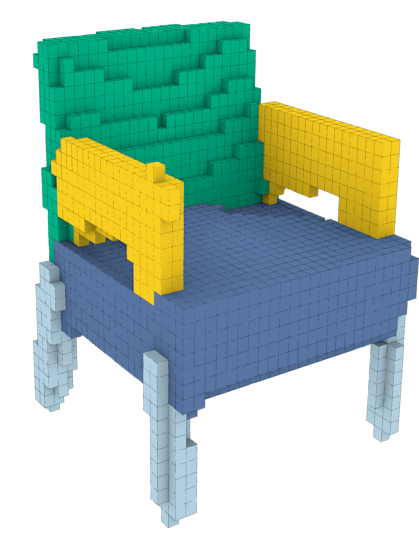}
\hfill
\includegraphics[width=0.15\linewidth]{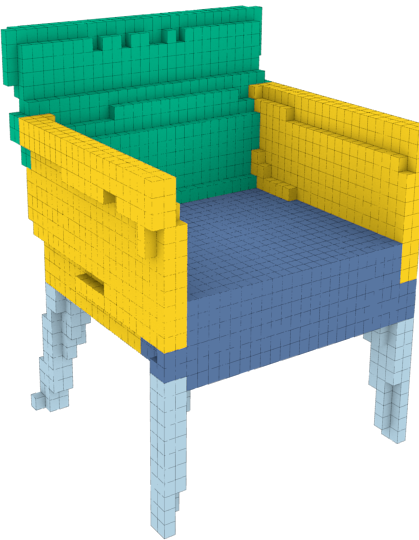}
\\
\rotatebox{90}{\hspace{1.5em}Back}
\hfill
\includegraphics[width=0.14\linewidth]{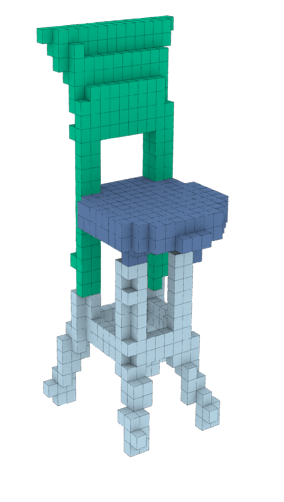}
\hfill
\includegraphics[width=0.14\linewidth]{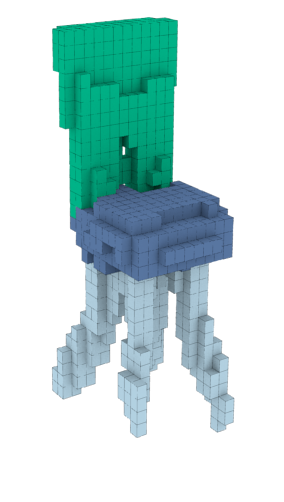}
\hfill
\includegraphics[width=0.14\linewidth]{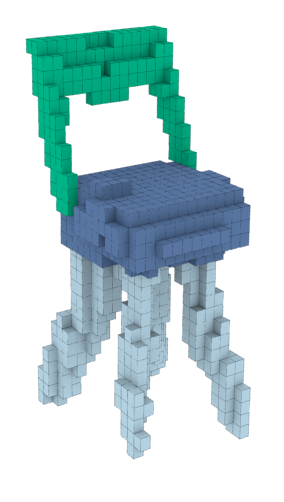}
\hfill\hfill
\includegraphics[width=0.14\linewidth]{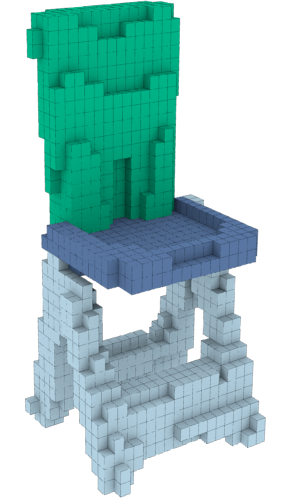}
\hfill
\includegraphics[width=0.14\linewidth]{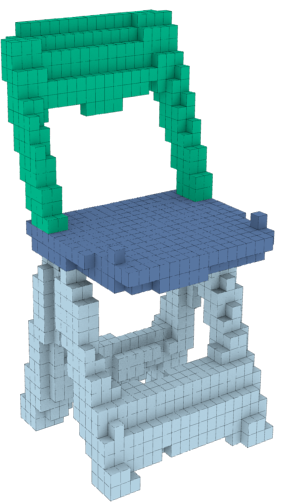}
\hfill
\includegraphics[width=0.14\linewidth]{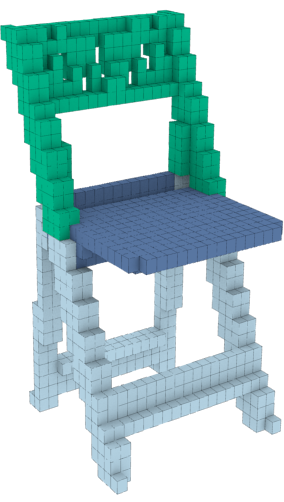}
\\[0.2em]
\begin{minipage}{.05\linewidth} \centerline{ } \end{minipage}
\hfill
\begin{minipage}{.15\linewidth}
\centerline{{\small GT$_1$}}
\end{minipage}%
\hfill
\begin{minipage}{.15\linewidth}
\centerline{{\small REC$_1$}}
\end{minipage}%
\hfill\hfill
\begin{minipage}{.15\linewidth}
\centerline{{\small SWAP$_1$}}
\end{minipage}%
\hfill
\begin{minipage}{.15\linewidth}
\centerline{{\small SWAP$_2$}}
\end{minipage}%
\hfill\hfill
\begin{minipage}{.15\linewidth}
\centerline{{\small REC$_2$}}
\end{minipage}%
\hfill
\begin{minipage}{.15\linewidth}
\centerline{{\small GT$_2$}}
\end{minipage}%
\\
\caption{Single part exchange experiment. GT$_{1/2}$ denote ground truth shapes, REC$_{1/2}$ - reconstruction results, SWAP$_{1/2}$ - part exchange results. Unlabeled shapes were used as an input.\label{fig:swap}}
\end{center}
\vspace{-1em}
\end{figure}

\paragraph{Shape composition by part exchange}
In this experiment, we used our structured latent space to randomly swap corresponding embedding coordinates of pairs of input shapes (\eg, embedding coordinates of legs or seats of two chairs), and reconstruct the new shapes using the Composer. The results are shown in Figure~\ref{fig:swap}, and demonstrate the ability of our system to perform accurate part exchange, while deforming the geometry of both the new and the existing parts to obtain a plausible result. See the supplementary material for additional results using four shape classes.

\begin{figure}[t]
\begin{center}
    \begin{minipage}{0.02\linewidth}
    \rotatebox{90}{\hspace{1em}GT}
    \\[0.1em]
    \rotatebox{90}{\hspace{-1em}Composed}
    \end{minipage}
    \hfill\hfill\hfill
    \begin{minipage}{0.6\linewidth}
    \includegraphics[width=0.23\textwidth]{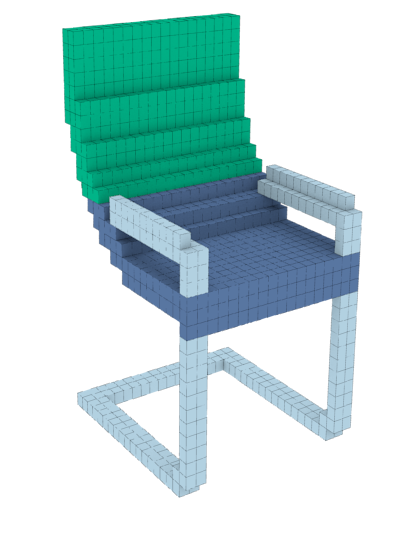}
    \hfill
	\includegraphics[width=0.23\textwidth]{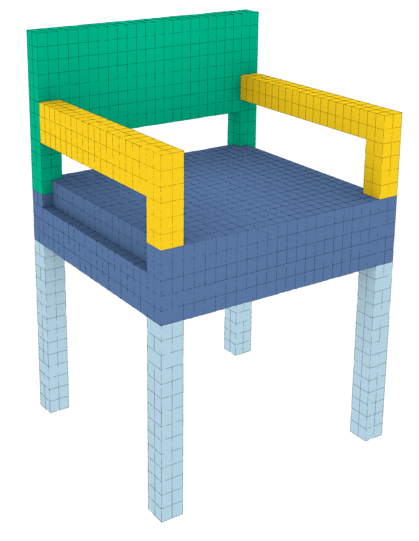}
    \hfill
    \includegraphics[width=0.23\textwidth]{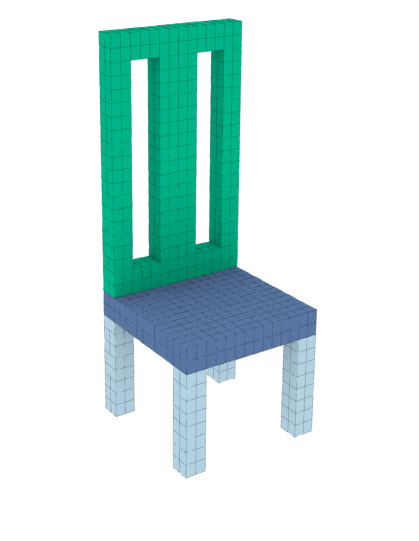}
    \hfill
	\includegraphics[width=0.23\textwidth]{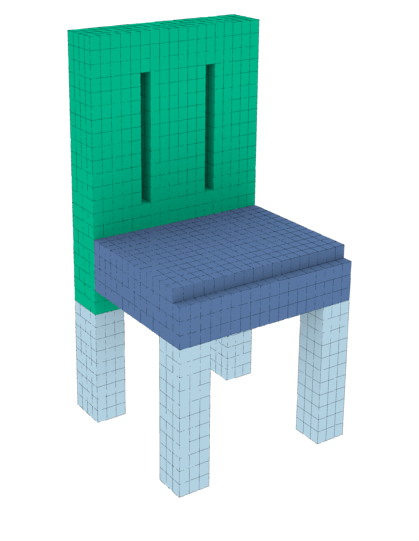}
    \\
    \includegraphics[width=0.23\textwidth]{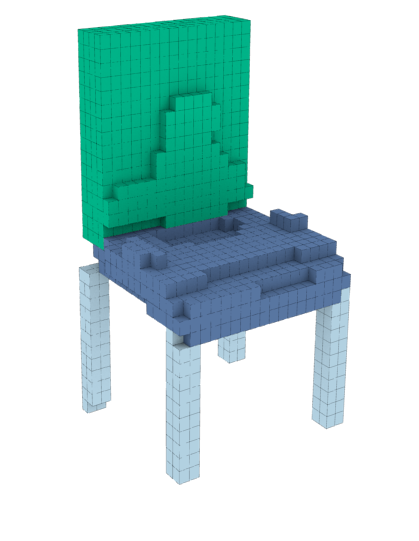}
    \hfill
	\includegraphics[width=0.23\textwidth]{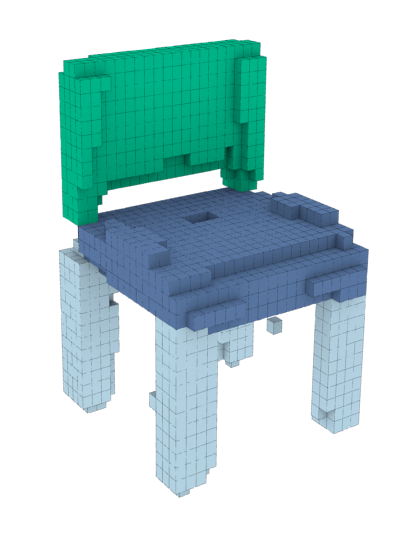}
    \hfill
    \includegraphics[width=0.23\textwidth]{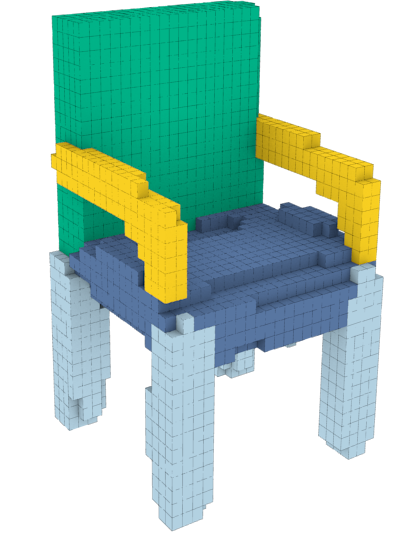}
    \hfill
	\includegraphics[width=0.23\textwidth]{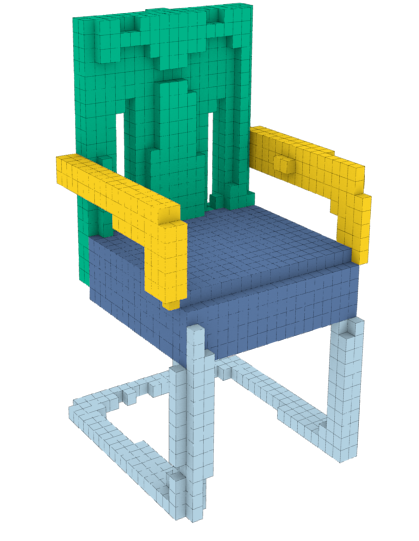}
    \end{minipage}
    \hfill \hfill
    \begin{minipage}{0.35\linewidth}
    \includegraphics[width=0.48\textwidth]{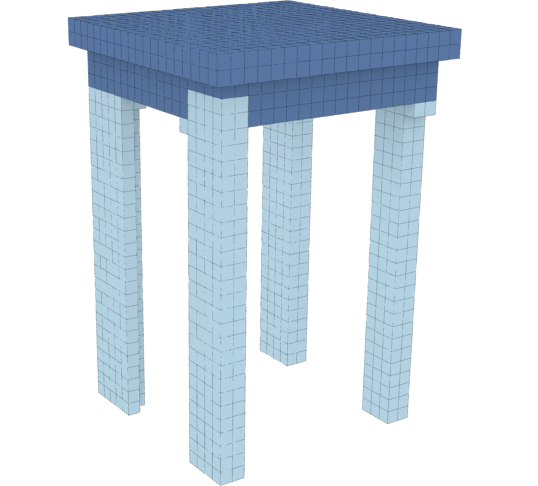}
    \hfill
	\includegraphics[width=0.48\textwidth]{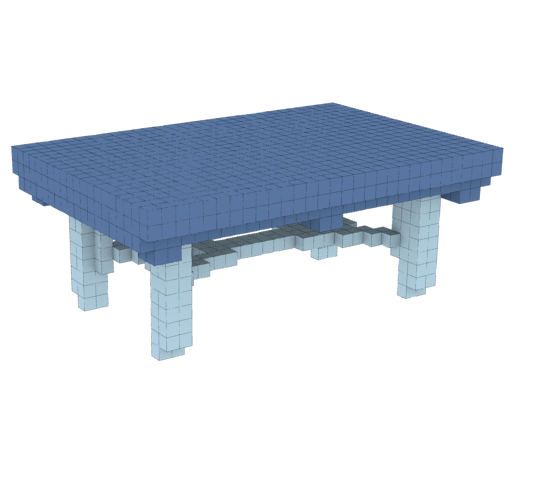}
    \\
    \includegraphics[width=0.48\textwidth]{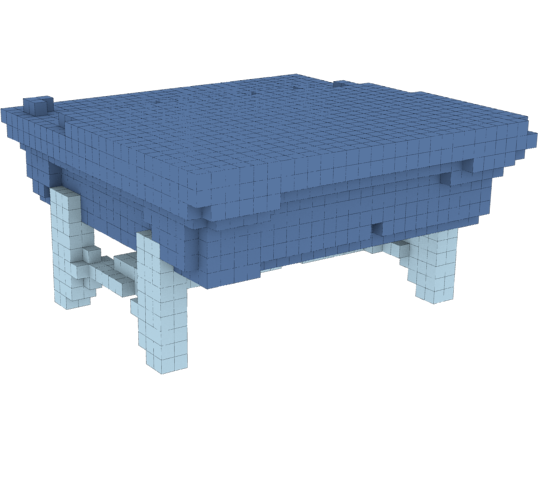}
    \hfill
	\includegraphics[width=0.48\textwidth]{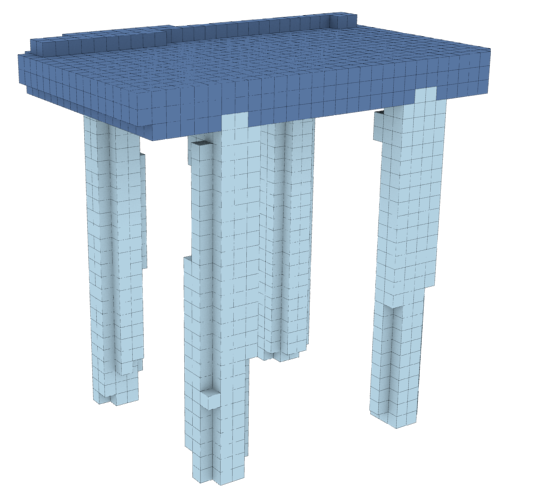}
    \end{minipage}
    \\
    \caption{Shape composition by random part assembly. The top row shows the ground truth (GT) shapes, and the bottom row - shapes assembled using the proposed approach (see Section~\ref{sec:random_compose}). Unlabeled shapes were used as an input.}
\label{fig:collect_chair}
\end{center}
\vspace{-1em}
\end{figure}

\vspace{-1em}
\paragraph{Shape composition by random part assembly\label{sec:random_compose}}
In this experiment we tested the ability of the proposed network to assemble shapes from random parts using our factorized latent space. Specifically, we mapped batches of input shapes into the latent space using the Decomposer, and created new shapes by randomly mixing the part embedding coordinates of the shapes in the batch, and reconstructing new shapes using the Composer. The results are shown in Figure~\ref{fig:collect_chair}, for chairs and tables, and illustrate the ability of the proposed method to combine parts from different shapes, scale and translate them so that the resulting shape looks realistic. See the supplementary material for additional shape composition results.

\vspace{-1em}
\paragraph{Full and partial interpolation in the embedding space}
In this experiment, we tested reconstruction from linearly interpolated embedding coordinates of complete shapes, as well as of a single semantic part. For the latter, we performed the part exchange experiment, described above, and interpolated the coordinates of that part, while keeping the rest of part embedding coordinates fixed. The results are shown in Figure~\ref{fig:full_ae_interp}. See the supplementary material additional interpolation results.  

\begin{figure*}[thb]
\begin{center}
    \rotatebox{90}{\hspace{1.7em}Whole}
	\hfill
	\includegraphics[width=0.07\textwidth]{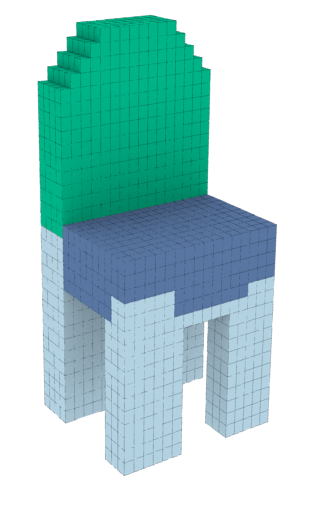}
    \hfill \hfill
	\includegraphics[width=0.07\textwidth]{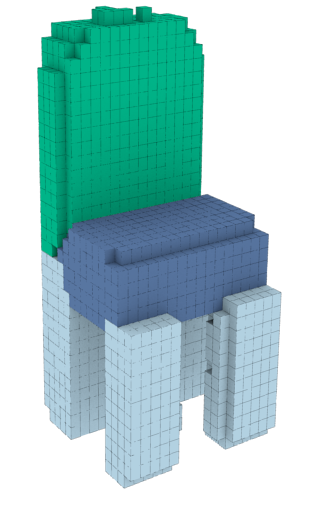}
    \hfill
	\includegraphics[width=0.07\textwidth]{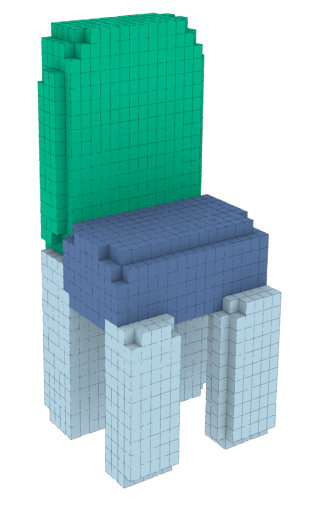}
    \hfill
	\includegraphics[width=0.07\textwidth]{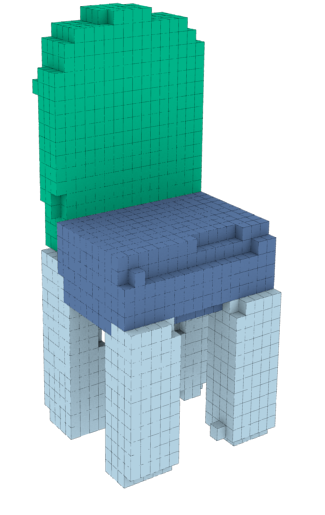}
    \hfill
	\includegraphics[width=0.07\textwidth]{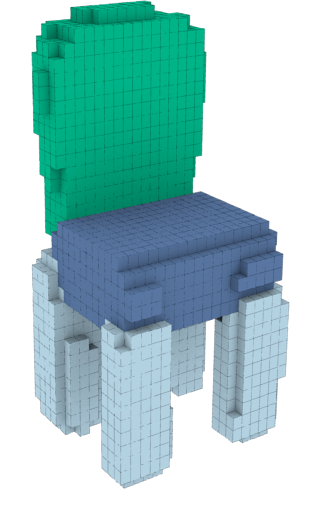}
    \hfill
	\includegraphics[width=0.07\textwidth]{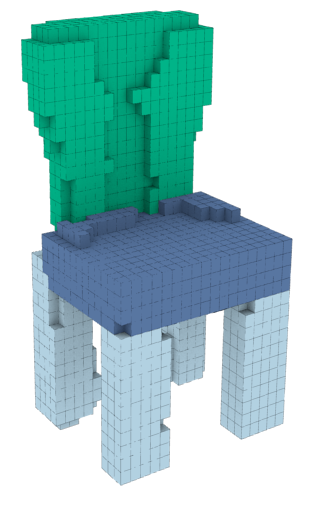}
    \hfill
 	\includegraphics[width=0.07\textwidth]{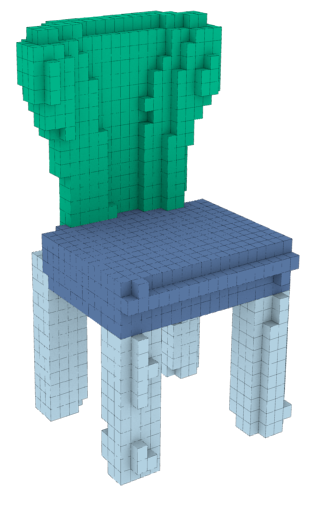}
    \hfill
	\includegraphics[width=0.07\textwidth]{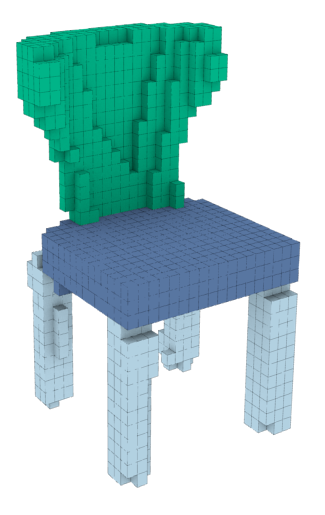}
    \hfill
 	\includegraphics[width=0.07\textwidth]{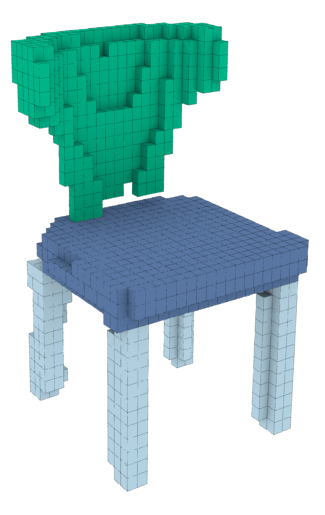}
    \hfill
	\includegraphics[width=0.07\textwidth]{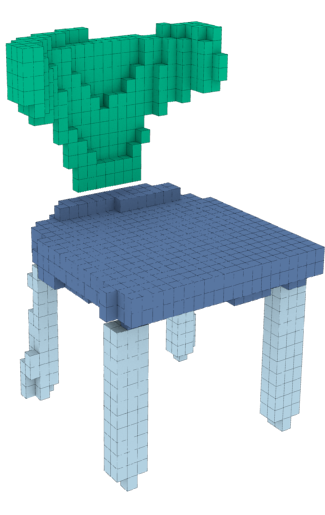}
    \hfill
	\includegraphics[width=0.07\textwidth]{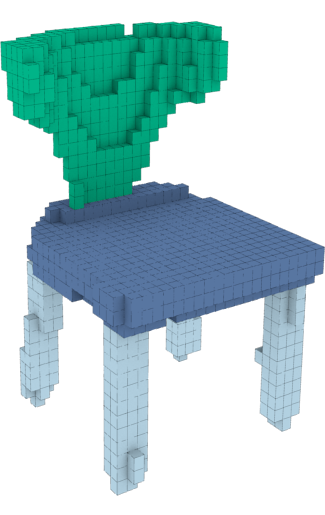}
    \hfill \hfill
	\includegraphics[width=0.07\textwidth]{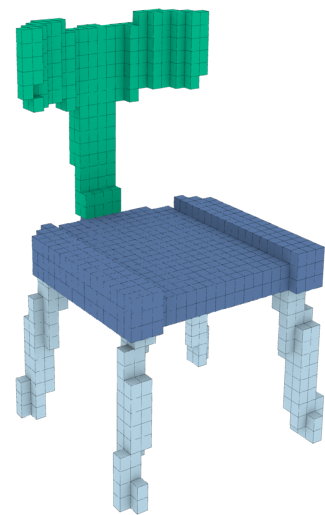}
	\\[0.1em]
    \rotatebox{90}{\hspace{2em}Back}
	\hfill
	\includegraphics[width=0.07\textwidth]{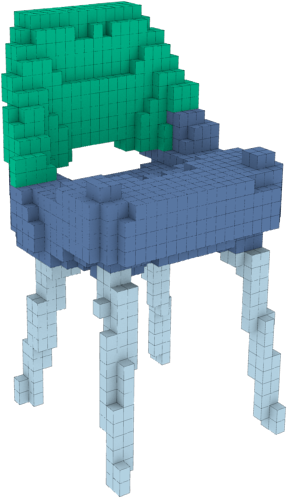}
    \hfill \hfill
	\includegraphics[width=0.07\textwidth]{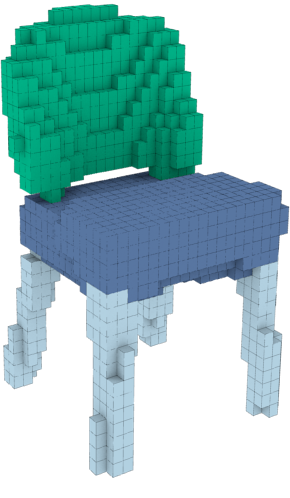}
	\hfill
	\includegraphics[width=0.07\textwidth]{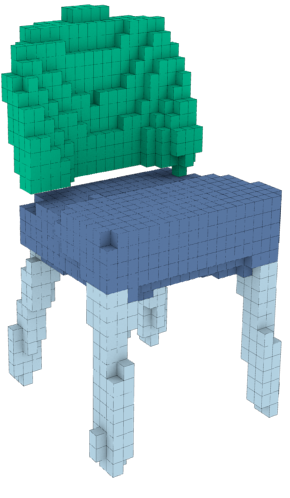}
    \hfill
	\includegraphics[width=0.07\textwidth]{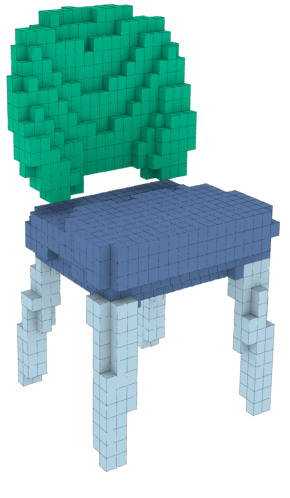}
    \hfill
	\includegraphics[width=0.07\textwidth]{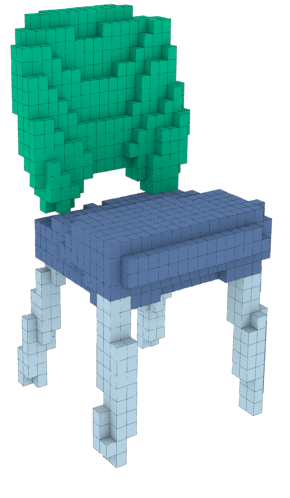}
    \hfill
	\includegraphics[width=0.07\textwidth]{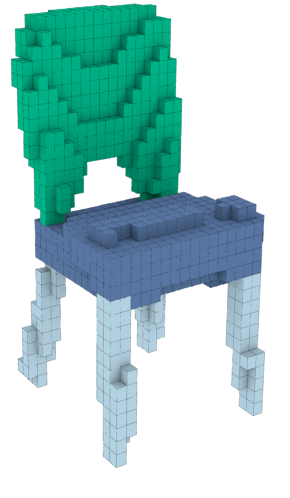}
    \hfill
 	\includegraphics[width=0.07\textwidth]{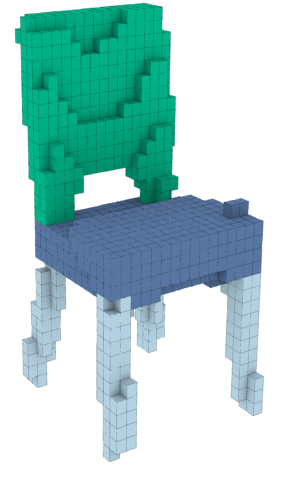}
    \hfill
	\includegraphics[width=0.07\textwidth]{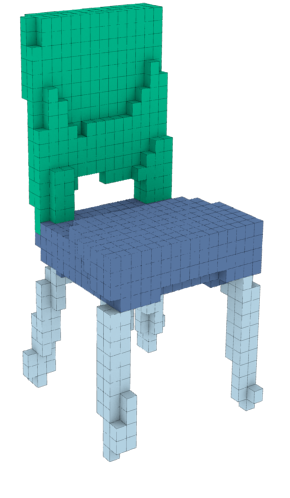}
    \hfill
 	\includegraphics[width=0.07\textwidth]{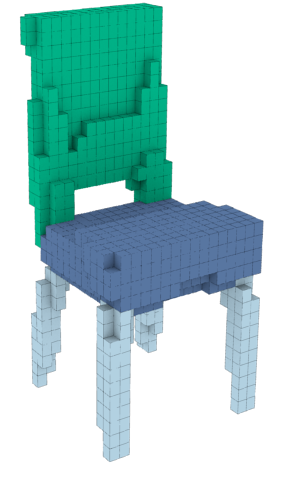}
    \hfill
	\includegraphics[width=0.07\textwidth]{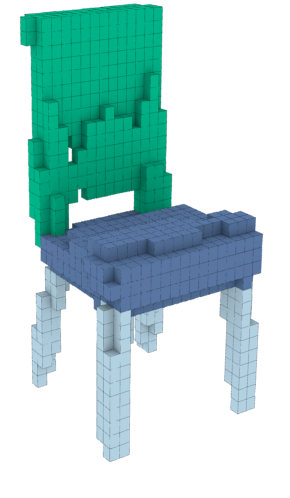}
    \hfill
	\includegraphics[width=0.07\textwidth]{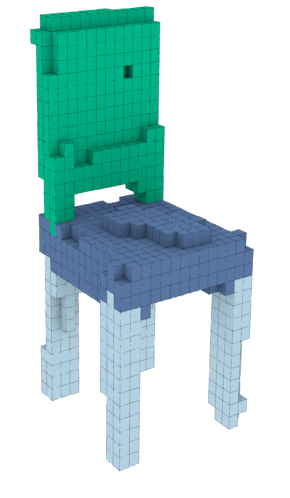}
    \hfill \hfill
	\includegraphics[width=0.07\textwidth]{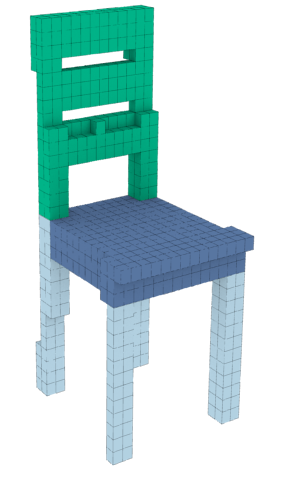}
    \\
    \begin{minipage}{0.03\linewidth} \centerline{ } \end{minipage}%
    \begin{minipage}{0.07\linewidth} \centerline{{\small GT$_1$}} \end{minipage}%
    \hfill \hfill
    \begin{minipage}{0.07\linewidth} \centerline{{\small REC$_1$}} \end{minipage}%
    \hfill
    \begin{minipage}{0.07\linewidth} \centerline{{\small $\alpha=\frac{1}{9}$}} \end{minipage}%
    \hfill
    \begin{minipage}{0.07\linewidth} \centerline{{\small $\alpha=\frac{2}{9}$}} \end{minipage}%
    \hfill
    \begin{minipage}{0.07\linewidth} \centerline{{\small $\alpha=\frac{3}{9}$}} \end{minipage}%
    \hfill
    \begin{minipage}{0.07\linewidth} \centerline{{\small $\alpha=\frac{4}{9}$}} \end{minipage}%
    \hfill
    \begin{minipage}{0.07\linewidth} \centerline{{\small $\alpha=\frac{5}{9}$}} \end{minipage}%
    \hfill
    \begin{minipage}{0.07\linewidth} \centerline{{\small $\alpha=\frac{6}{9}$}} \end{minipage}%
    \hfill
    \begin{minipage}{0.07\linewidth} \centerline{{\small $\alpha=\frac{7}{9}$}} \end{minipage}%
    \hfill
    \begin{minipage}{0.07\linewidth} \centerline{{\small $\alpha=\frac{8}{9}$}} \end{minipage}%
    \hfill
    \begin{minipage}{0.07\linewidth} \centerline{{\small REC$_2$}} \end{minipage}%
    \hfill \hfill
    \begin{minipage}{0.07\linewidth} \centerline{{\small GT$_2$}} \end{minipage}%
\caption{Example of a whole (top) and partial (bottom) shape interpolation. GT$_{1/2}$ denote original models, REC$_{1/2}$ - their reconstructions, and linear interpolation results are in the middle. Unlabeled shapes were used as an input.}
\label{fig:full_ae_interp}
\end{center}
\vspace{-0.7em}
\end{figure*}

\subsection{Latent space and projection matrix analysis}
The latent space obtained using the proposed method exhibits clear separation into subspaces corresponding to different semantic parts. The projection matrices, while not not being strictly orthogonal, as required for the partition of the identity (\ref{eq:part_of_identity}), have low effective ranks, which is in line with the clear separation into non-overlapping subspaces produced by them. See the supplementary material for the latent space and the projection matrices visualization.

\begin{table*}[tb]
\begin{center}
\begin{tabular}{|l||c|c|c|c|c|c|c|c|c|c|c|c|c|c|c|}
\hline
\backslashbox{Method}{\begin{tabular}{@{}c@{}}Metric\end{tabular}}
& mIoU & \begin{tabular}{@{}c@{}}mIoU\\(parts)\end{tabular} &
\multicolumn{3}{c|}{Connectivity} & \multicolumn{3}{c|}{\begin{tabular}{@{}c@{}}Classifier\\accuracy\end{tabular}} & \multicolumn{3}{c|}{\begin{tabular}{@{}c@{}}Symmetry\\score\end{tabular}} \\
\cline{2-12}
& Rec. & Rec. & Rec. & Swap & Mix & Rec. & Swap & Mix & Rec. & Swap & Mix  \\
\hline\hline
Our method & 0.64 & 0.65 & \bf{0.82} & \bf{0.71} & \bf{0.65} & \bf{0.95} & 0.89 & \bf{0.83} & 0.95 & 0.95 & 0.95 \\ 
\Xhline{2\arrayrulewidth}
W/o cycle loss & 0.63 & 0.66 & 0.74 & 0.62 & 0.54 & 0.93 & 0.84 & 0.80 & 0.96 & 0.96 & 0.95 \\ 
\hline
Fixed projection & 0.63 & 0.65 & 0.72 & 0.61 & 0.58 & 0.94 & 0.86 & 0.77 & 0.94 & 0.95 & 0.95 \\ 
\hline
Composer w/o STN & \bf{0.75} & \bf{0.8} & 0.69 & 0.48 & 0.23 & \bf{0.95} & \bf{0.9} & 0.71 & 0.95 & 0.91 & 0.85 \\ 
\Xhline{2\arrayrulewidth}
Naive placement & - & - & - & 0.68 & 0.62 & 0.61 & 0.47 & 0.21 & - & \bf{0.96} & \bf{0.96} \\ 
\hline
ComplementMe
& - & - & - & \bf{0.71} & 0.47 & - & 0.66 & 0.43 & - & 0.66 & 0.43 \\ 
\hline
Segmentation+STN & - & - & - & 0.41 & 0.64 & - & 0.64 & 0.36 & - & 0.77 & 0.77 \\
\hline
\end{tabular}
\end{center}
\vspace{-1em}
\caption{Ablation study results. The evaluation metrics are mean Intersection over Union (\emph{mIoU}), per-part mean IoU (\emph{mIoU (parts)}), shape \emph{connectivity} measure, binary shape \emph{classifier accuracy}, and shape \emph{symmetry score}. Rec., Swap and Mix stand for the shape reconstruction, part exchange and random part assembly experiment results, respectively (see Section~\ref{sec:composite_synthesis}).
See Section~\ref{sec:ablation} for a detailed description of the compared methods and the evaluation metrics. \label{tbl:ablation_eval}}
\end{table*}


\subsection{Ablation study and comparison with existing approaches}
\label{sec:ablation}
\subsubsection{Ablation study}
To highlight the importance of the different elements of our approach, we conducted an ablation study, where we used several variants of the proposed method, listed below.

\vspace{-1em}
\paragraph{Fixed projection matrices} Instead of using learned projection matrices in the Decomposer, the $n$-dimensional shape encoding is split into $K$ consecutive equal-sized segments, which correspond to different part embedding subspaces. This is equivalent to using constant projection matrices, where the elements of the rows corresponding to a particular embedding space dimensions are $1$, and the rest of the elements are $0$. 

\vspace{-1em}
\paragraph{Composer without STN} We substituted the proposed composer, consisting of the part decoder and the STN, with a single decoder producing a labeled shape. The decoder receives the sum of part encodings as an input, processes it with two FC layers to combine information from different parts, and then reconstructs a shape with parts labels using a series of deconvolution steps, similar to the part decoder in the proposed architecture.



\vspace{-1em}
\paragraph{Without cycle loss} We removed the cycle loss component during the network training.


\subsubsection{Comparison with existing methods}
Most existing methods for composite shape modeling operate on triangulated meshes with precise part segmentation. Hence, they are not directly applicable to the large-scale ShapeNet dataset with less precise segmentation, preventing a fair comparison. 
We therefore added the following comparisons with modern neural-net-based techniques: we combined the state-of-the-art ComplementMe method \cite{sung2017complementme} with a 3D-CNN segmentation network  \cite{qi2017pointnet}. From the former we used the \emph{component placement network}, which, given a partial shape and a complementary component, produces a 3-D translation to place the component correctly w.r.t. the partial shape. To produce the "to-be-added" component we used a 3D-CNN segmentation network, described in \cite{qi2017pointnet}, which achieved a state-of-the-art mean Intersection over Union (mIoU) of 0.91 on the test set. Together, these two networks replace our proposed Decomposer-Composer. Both networks were trained using the same training data as the proposed method. 
This method is denoted by \emph{ComplementMe} in Table~\ref{tbl:ablation_eval}.

For an additional comparison, instead of the placement network of ComplementMe we utilized the spatial transformer network. Here, the STN was trained using the ground truth shape parts, and at test time it was applied to the results of the segmentation network, described above. This method is denoted by \emph{Segmentation+STN} in Table~\ref{tbl:ablation_eval}.

Finally, we compared the proposed method to a baseline shape composition network. Given ground-truth shape parts, it composes new shapes from these parts by placing them at their original locations in the source shapes they were extracted from. All the shapes in our dataset are centered and uniformly scaled to fill the unit volume, and there exist clusters of geometrically and semantically similar shapes. Thus, we can expect that even this naive approach without part transformations will produce plausible results in some cases. This method is denoted by \emph{Naive placement} in Table~\ref{tbl:ablation_eval}.

See the supplementary material for an additional qualitative comparison with 3D-GAN \cite{3dgan} and G2LGAN \cite{G2L18}, using $64\times64\times 64$ voxelized shapes.

\subsubsection{Evaluation metrics\label{sec:metrics}}
\paragraph{Mean Intersection over Union (mIoU)} is commonly used to evaluate the performance of segmentation algorithms~\cite{long2015fully}. Here, we use it as a metric for the reconstruction quality. We computed the mIoU for both actual-sized reconstructed parts, and scaled and centered parts (when applicable). We denote the two measures by \emph{mIoU} and \emph{mIoU (parts)} in Table~\ref{tbl:ablation_eval}.

\vspace{-1em}\paragraph{Connectivity} In part based shape synthesis, one pathological issue is that parts are often disconnected, or penetrate each other. 
Here, we would like to benchmark the quality of part placement, in terms of part connectivity. For each $32\times 32\times 32$ volume, we compute the frequency of the shape forming a single connected component, and report it as \emph{Connectivity} in Table~\ref{tbl:ablation_eval}.

\vspace{-1em}\paragraph{Classification accuracy} To measure the shape composition quality of different methods, we trained a binary neural classifier to distinguish between ground-truth whole chairs (acting as positive examples) and chairs produced by naively placing random chair parts together (acting as negative examples). To construct the negative examples, we randomly combined ground-truth shape parts, by adding a certain semantic part only once, and placing the parts at their original locations in the source shapes they were extracted from. In addition, we removed negative examples assembled from parts from geometrically and semantically similar chairs, since such part arrangement could produce plausible shapes incorrectly placed in the negative example set. The attained classification accuracy on the test set was $\sim~88\%$. For a given set of chairs, we report the average classification score. Details of the network can be found in the supplementary material. The results are reported as \emph{Classifier accuracy} in Table~\ref{tbl:ablation_eval}.

    
    

\vspace{-1em}\paragraph{Symmetry} The chair shapes in the ShapeNet are predominantly bilaterally symmetric, with vertical symmetry plane. Thus, similar to \cite{G2L18}, we evaluated the symmetry of the reconstructed shapes, and defined the \emph{Symmetry score} as the percentage of the matched voxels (filled or empty) in the reconstructed volume and its reflection with respect to the vertical symmetry plane. We performed this evaluation using binarized reconstruction results, effectively measuring the global symmetry of the shapes. For the evaluation, we used the shapes in the test set (690 shapes), and conducted three types of experiments: shape reconstruction, single random part exchange between a pair of random shapes, shape composition by random part assembly. The experiments are described in more detail in Sections~\ref{sec:reconstruction} and \ref{sec:composite_synthesis}.

\subsubsection{Evaluation result discussion}
According to all metrics, our method outperforms or performs on par with all the baselines, and \emph{significantly} outperforms other existing methods. This shows that our design choices - the cycle loss, learned projection matrices and usage of the STN, help to achieve plausible results both when reconstructing shapes, and when performing composite shape synthesis. This is especially pronounced in the connectivity test results, illustrating that these design choices are necessary for achieving good assembly quality.

In the classifier accuracy test and the symmetry test, the proposed method performs slightly better or on par with all baselines considered in the ablation study. It seems that both these tests are less sensitive to disconnected shape components, and most advantage that the proposed method achieves over the baselines is in its composition robustness. As expected, the naive placement also achieves high symmetry score, since it preserves the symmetry of the ground-truth parts during shape assembly.

According to the mIoU and per-part mIoU metrics, the proposed method performs on par with all baselines, except when using the simple version of the Composer, without STN. This follows from the fact that the proposed system, while reconstructing better fine geometry features, decomposes the problem into two inference problems, for the geometry and the transformation, and thus does not produce as faithful reconstruction of the original model as the simple decoder. Notably, this version of the architecture achieves worst connectivity scores for all compared methods, which follows from the fact that such a Decomposer is unable to faithfully reconstruct fine shape details. Please see the supplementary material for a qualitative comparison of the results of all the compared methods.


\section{Conclusions and future work}
\label{seq:conslusions}

We presented a Decomposer-Composer network for structure-aware 3D shape modelling. It is able to generate a factorized latent shape representation, where different semantic part embedding coordinates lie in separate linear subspaces. The subspace factorization allows us to perform shape manipulation via part embedding coordinates, exchange parts between shapes, or synthesize novel shapes by assembling a shape from random parts. Qualitative results show that the proposed system can generate high fidelity 3D shapes and meaningful part manipulations. Quantitative results shows we are competitive in the mIOU, connectivity, symmetry and classification benchmarks.

While the proposed approach makes a step toward automatic shape-from-part assembly, it has several limitations. First, while we can generate high-fidelity shapes at a relatively low resolution, memory limitations do not allow us to work with voxelized shapes of higher resolution. Memory-efficient architectures, such as OctNet~\cite{riegler2017octnet} and PointGrid~\cite{le2018pointgrid}, may help alleviate this constraint. Alternatively, using point-based shape representations and compatible deep network architectures, such as~\cite{qi2017pointnet}, may also reduce the memory requirements and increase the output resolution.

Secondly, we made a simplifying assumption that a plausible shape can be assembled from parts using per-part affine transformations, which represent only a subset of possible transformations. While this assumption simplifies the training, it is quite restrictive in terms of the deformations we can perform. In future work, we will consider general transformations which have higher degree of freedom, such as a 3D thin plate spline or a general deformation fields. 
To promote better part connectivity, we will explore additional shape connectivity preservation losses, similar to \cite{G2L18}.
Finally, we have been using a cross-entropy loss to measure the shape reconstruction quality; it would be interesting to investigate the use of a GAN-type loss in this structure-aware shape generation context.

{\small
\bibliographystyle{ieee_fullname}
\bibliography{sfp-iccv}
}

\end{document}